\documentclass[conference]{IEEEtran}

\usepackage{amsmath,amssymb,amsfonts}
\usepackage{algorithmic}
\usepackage{algorithm}
\usepackage{graphicx}
\usepackage{textcomp}
\usepackage{xcolor}
\usepackage{hyperref}
\usepackage{booktabs}
\usepackage{multirow}
\usepackage{subcaption}
\graphicspath{{./}}

\begin{document}

\title{Binary Road Surface Classification Using Machine Learning on Production Vehicle Signals During Cruising}

\author{
\IEEEauthorblockN{Vishal Hariharan}
\IEEEauthorblockA{Global Tire Intelligence and Solutions\\
The Goodyear Tire and Rubber Company\\
Akron, Ohio, USA\\
hvishal512@gmail.com}
\and
\IEEEauthorblockN{Salar Basiri}
\IEEEauthorblockA{Department of Mechanical Engg.\\
University of Illinois\\
Urbana Champaign (UIUC), USA\\
sbasiri2@illinois.edu}
\and
\IEEEauthorblockN{Kanwar Bharat Singh}
\IEEEauthorblockA{Global Tire Intelligence and Solutions\\
The Goodyear Tire and Rubber Company\\
Akron, Ohio, USA\\
kanwar-bharat\_singh@goodyear.com}
}

\maketitle

\begin{abstract}
Knowledge of real-time road slipperiness — or even better, a refined estimate of peak grip potential — is a critical input for vehicle warning and intervention control systems. Typically, friction is estimated through dynamics based recursive estimators by calculating the slip slope, however, its efficacy is heavily constrained by the vehicle dynamic scenario. When the vehicle is cruising and there is little to no slip, these methods become ineffective due to the inability of present-day production grade sensors (like wheel speed sensors) and methods in either measuring or accurately estimating micro slip which ends up being crucial in distinguishing different surfaces. To address this challenge, correlation between vehicle signals and road surface condition during cruising needs to be uncovered using machine learning. In this paper, a feature-based framework and an end-to-end data-driven framework are used to correlate the statistics of vehicle dynamics behavior to the condition of the road surface and perform binary classification into grip (dry/damp) or slip (snow/ice) condition. A sliding window approach is adopted to batch a short buffered window of wheel speeds, wheel torques, longitudinal acceleration, steering angle and yaw rate which are fed into a machine learning module for predicting the road state. Validation results on public road data show scenarios where the data driven method identifies the road surface correctly even during cruising, showing promise for accurate data driven friction-related state estimators in the field of tire and vehicle dynamics.
\end{abstract}


\section{Introduction}

Detecting road surface conditions is essential for accurate tire road friction estimation, a critical factor in vehicle safety applications such as collision warning, adaptive cruise control, and autonomous emergency braking (AEB) \cite{Du2024}. Time to collision (TTC) in present day AEB systems is calculated based on ego vehicle speed, speed of vehicle in front, and distance between the two vehicles \cite{Yang2022}. However, the calculation is done assuming that the friction value is that of a dry surface to avoid false positives. Having a reliable friction estimator will not only eliminate false positives but provide the vehicle additional time to perform control actions like applying the brakes earlier and adjusting wheel torques to avoid imminent collision on slippery surfaces such as snow and ice \cite{Yi2002, Kim2018}.

\subsection{Existing approaches and drawbacks}

Correlating tire stiffness obtained through slip slope state estimators with different road conditions is a popular approach, although it is controversial as to whether there is a clear correlation between tire stiffness and tire road grip level due to factors like temperature, wear level, inflation pressure, etc \cite{Acosta2017}. Additionally, when the vehicle is in cruising state, the estimator uncertainty increases \cite{Gustafsson1997-wq}. Methods based on Recursive Least Squares (RLS) have been used to identify tire-road friction in real time \cite{MooryongChoi2013, Zhao2014}, and fusion of model-based and data-based approaches has been explored to improve robustness \cite{Tang2023}. To overcome the limitation of lack of clear friction signature in tire stiffness, slip ratio derivative can be used to regress the grip, eliminating dependency on force estimators. Although a second-order term like the slip ratio derivative might contain information about different road surfaces, it still relies on the assumption of having a robust vehicle velocity estimator in all driving conditions. In theory, a data driven method should be able to extract information directly from onboard signals, as vehicle velocity and slip ratio derivative can both be derived from onboard signals. Onboard signals addressed in this work refers to wheel speeds, vehicle level accelerations, yaw rate, steering angle, and wheel torques. Recently, a promising approach that involves extrapolating the tire force slip curve using Bayesian methods \cite{Berntorp2019} to estimate the peak grip with uncertainty bounds has gained attraction. Model learning of the friction-slip dependency under standard driving conditions has also been explored \cite{Mussot2022}, though primarily validated on controlled test data rather than diverse public road conditions. Data-driven friction estimation using onboard signals with uncertainty evaluation \cite{Chen2025} has shown promise, but the question of whether surface state can be reliably classified from cruising data alone remains open.

\subsection{Our contribution}

This work demonstrates that grip/slip road-state classification during low-excitation cruising can be performed using only production-feasible onboard vehicle signals. Two complementary approaches are evaluated: a physically motivated feature-based XGBoost model and an end-to-end 1D CNN operating on short multichannel time-series buffers. The study compares model performance, feature/channel importance, qualitative unseen-run behavior, and real-time embedded feasibility, highlighting both the promise and limitations of data-driven road slipperiness detection under cruising conditions.

\section{Methodology}

\subsection{Dataset and Pre-processing}

The dataset used in this work consists of close to 10 hours of snow (includes ice) and 20 hours of dry (includes damp) surfaces mainly during mild driving conditions on public roads with a few slightly aggressive maneuvers on test tracks collected using an all-wheel-drive EV platform equipped with winter tires. Input signals are sampled at 100 Hz to capture vehicle response in different conditions on the road surface \cite{Acosta2017, Chen2025}. One of the challenges associated with training ML models is obtaining high-quality labeled data. To automatically label vehicle onboard signals, an internal optical sensor was used to obtain road conditions such as dry, moist, wet, pool, snow, or ice to label time series buffers of onboard signals such as grip (dry/moist) or slip (snow/ice). Wet condition detection is out of the scope of this study as the slipperiness signature can depend on the thickness of the water film and there was no real way to obtain the measurement of the thickness of the water. Another reason to exclude wet class is the similar tire behavior on dry and wet conditions in the linear regime of tire force slip curve \cite{Claeys2001}. An advantage with training Machine Learning models on physical signals such as that arising from a vehicle is the ease with which min max normalization can be applied to bound all the signals to be between 0 and 1 to stabilize model training. The min and max values for each channel are set apriori on the basis of what is physically possible. It is important to note that the min and max values are set as global variables and do not depend on a particular instance. Instance-based normalization loses meaning of relative magnitude difference and hence is not appropriate for this type of time series data.

\subsection{Model architecture}

\subsubsection{Feature based}

\begin{itemize}
\item Ratio of driven vs non-driven wheel speeds
\item RMS of slip ratio
\item RMS of slip ratio derivative
\item RMS of longitudinal acceleration
\end{itemize}

The above features are selected based on \cite{Druta2023, Jang2020-hi, Hou2017-ar, Li2006-xv}. By using the best features reported in literature to detect road slipperiness during cruising conditions, this work examines their variation and interaction on a large-scale dataset obtained from public roads. In an all-wheel-drive vehicle, due to torque vectoring, the driven axle can change between acceleration and braking scenarios. However, for the platform used in this study the torque vectoring mechanism during cruising always sent the torques to the front wheels enabling to determine a driven axle. It is also easier to calculate vehicle speed and hence slip ratio during cruising. For the above reasons, the feature-based method involves constraining the dataset to process 4 second buffers obtained from 0.1 second stride length where the absolute value of median longitudinal acceleration is less than 0.05G m/s$^2$ (G $= 9.81$) along with other constraints such as maximum difference across the 4 wheels being less than 2 kph, yaw rate between $-5$ deg/s to $5$ deg/s, steering angle between $-10$ degrees to $10$ degrees and torques between $-1000$ Nm to $1000$ Nm. While each individual feature itself need not show stark difference between dry and snow class during cruising, it is possible to learn a non-linear discriminative function by using an ensemble method based on eXtreme Gradient Boosting (XGBoost) \cite{ChenGuestrin2016}.

\subsubsection{1D CNN}

Road friction signatures reflect as difference between wheel speeds and vehicle speed, as well as frequency content of wheel speeds and torques \cite{Umeno2002, Chen2017}. Frequency representations of signals however are expensive to compute in a real time setup, therefore a 1D Convolutional Neural Network \cite{Kiranyaz2021} is used to directly learn time based and implicit frequency based pattern differences from data. Features learned by the filters at different depths of the network should in theory capture frequency related phenomena. However, it is possible that the source signals contain noise emanating from the engine and transmission leading to model overfitting. For production settings, appropriate band pass or wavelet denoising \cite{Halidou2023, Donoho1995} should be applied to the signals before subjecting them to end to end model training. For the purposes of this work, the goal is to train a network as shown in Table~\ref{tab:cnn_architecture} that is robust to other noise sources given the scale of data.

The input consists of multi-channel time series buffers of 2 seconds with 0.1 seconds stride length containing wheel speeds, wheel torques, longitudinal acceleration, lateral acceleration, vertical acceleration, steering angle, and yaw rate (13 channels total). Input data instances are constrained to have median acceleration $\sqrt{a_x^2 + a_y^2}$ less than 0.1G, maximum wheel speed difference across the 4 wheels less than 5 kph, yaw rate between $-30$ deg/s and $30$ deg/s, and steering angle between $-45$ and $45$ degrees.

\begin{table}[t]
\caption{1D CNN Model Architecture}
\label{tab:cnn_architecture}
\centering
\small
\begin{tabular}{ll}
\toprule
\textbf{Stage} & \textbf{Layer Configuration} \\
\midrule
\multirow{4}{*}{Block 1} & Conv1d(in=13, out=32, kernel=3, stride=1, pad=1) \\
 & LayerNorm(32, 200) \\
 & ReLU() \\
 & MaxPool1d(kernel=2, stride=2) \\
\midrule
\multirow{4}{*}{Block 2} & Conv1d(in=32, out=64, kernel=3, stride=1, pad=1) \\
 & LayerNorm(64, 100) \\
 & ReLU() \\
 & MaxPool1d(kernel=2, stride=2) \\
\midrule
\multirow{3}{*}{Block 3} & Conv1d(in=64, out=128, kernel=3, stride=1, pad=1) \\
 & LayerNorm(128, 50) \\
 & ReLU() \\
\midrule
\multirow{2}{*}{Global Pool} & AdaptiveAvgPool1d(output\_size=1) \\
 & Flatten(start\_dim=1) \\
\midrule
\multirow{5}{*}{FC Head} & Linear(in=128, out=64) \\
 & LayerNorm(64) \\
 & ReLU() \\
 & Dropout(p=0.5) \\
 & Linear(in=64, out=1) $\rightarrow$ Sigmoid() \\
\bottomrule
\end{tabular}
\end{table}

\section{Model Performance and Discussion}

\subsection{Feature based model}

\begin{figure}[t]
\centering
\includegraphics[width=\columnwidth]{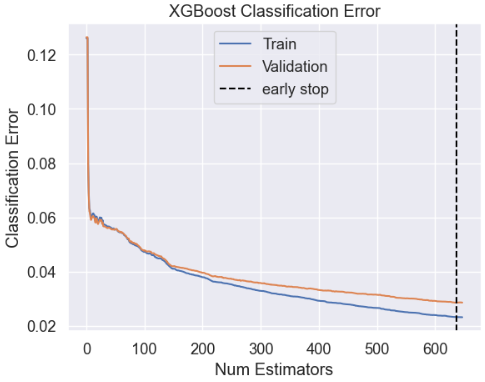}
\caption{Classification workflow.}
\label{fig:pipeline}
\end{figure}

\begin{figure}[t]
\centering
\includegraphics[width=\columnwidth]{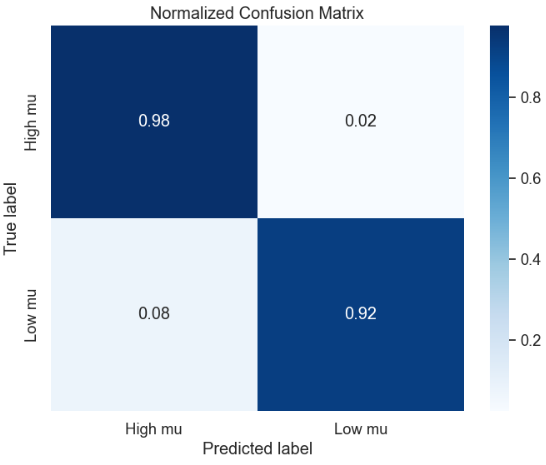}
\caption{Confusion matrix of tree-based model on test set.}
\label{fig:tree_cm}
\end{figure}

\begin{figure}[t]
\centering
\includegraphics[width=\columnwidth]{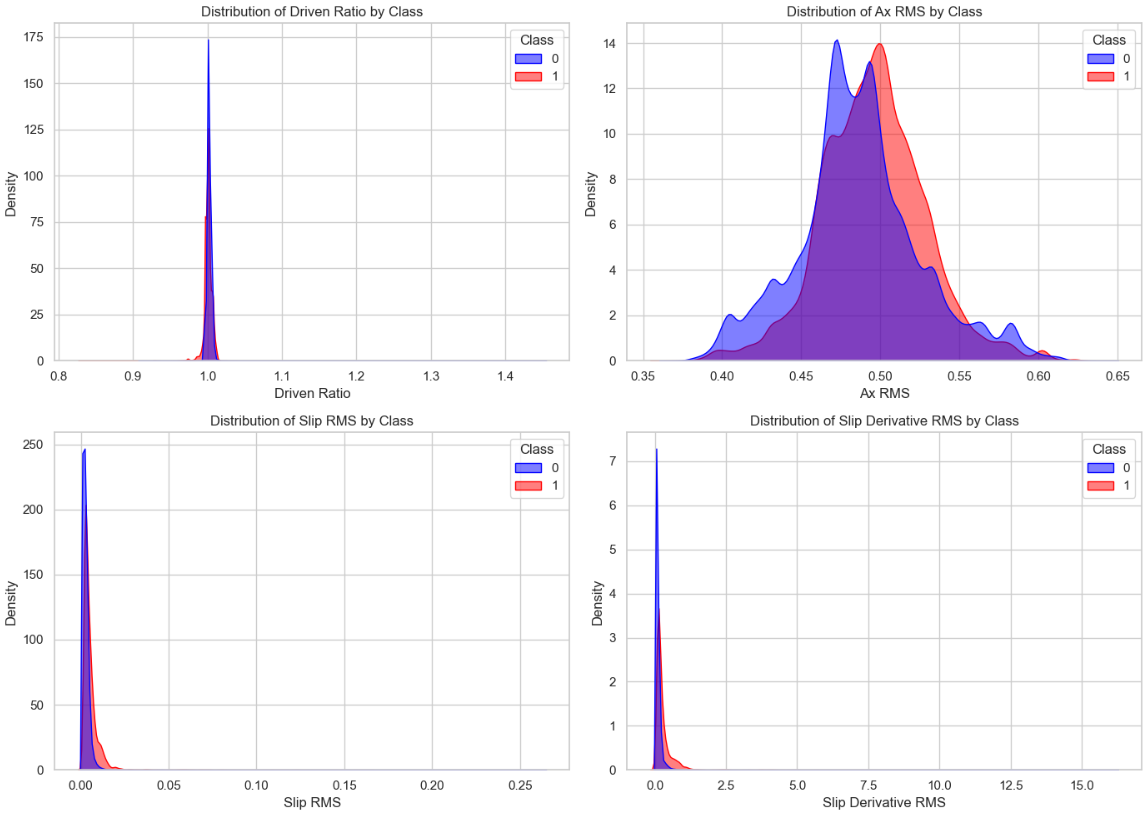}
\caption{Feature variation across classes dry (0) and snow (1).}
\label{fig:histograms}
\end{figure}

Training strategy includes assigning inverse square root class importance to account for the 2:1 class imbalance. A 0.7/0.15/0.15 train/validation/test split is applied at the individual window level. It should be noted that because a sliding window with 0.1\,s stride is used over continuous driving runs, adjacent windows in train and test sets may originate from the same driving run and share temporal overlap. The implications of this are discussed in Section~VII. Early stopping is used to halt learning when validation accuracy ceases to improve for a certain number of steps as shown in Fig.~\ref{fig:pipeline}. The tree-based model performance on the test set is shown in Fig.~\ref{fig:tree_cm}. The tree-based model achieves 98\% recall for the grip class (high $\mu$) and 92\% recall for the slip class (low $\mu$), with grip precision of 92.5\% and slip precision of 97.9\%. Histograms of different features extracted from 4 seconds of buffered data with 0.1 seconds stride length are shown in Fig.~\ref{fig:histograms}. Slip ratio derivative shows the highest variation between the grip (0) and slip (1) classes. While there is minimal variation of $a_x$ RMS and slip ratio of driven non-driven wheel speeds, it does not mean that they are not useful. In fact, looking at feature importance scores in Table~\ref{tab:tree_importance}, ratio of driven to non driven speeds serve as the second most important feature for the tree-based model to learn differentiating interaction terms.

\begin{table}[t]
\caption{Tree Feature Importance Scores}
\label{tab:tree_importance}
\centering
\begin{tabular}{lc}
\toprule
\textbf{Feature} & \textbf{Importance Score} \\
\midrule
Slip ratio derivative RMS & 0.4059 \\
Ratio of driven/non-driven speeds & 0.2748 \\
Slip ratio RMS & 0.1615 \\
Longitudinal Acceleration RMS & 0.1577 \\
\bottomrule
\end{tabular}
\end{table}

However, good performance on the confusion matrices alone does not guarantee generalization since a lot of the correctly classified instances could have been biased towards a certain vehicle dynamic scenario. To truly evaluate how applicable such a method is for production, one needs to evaluate performance on different unseen runs for longer periods. This is covered in the next section.

\subsection{1D CNN}

\begin{figure}[t]
\centering
\includegraphics[width=\columnwidth]{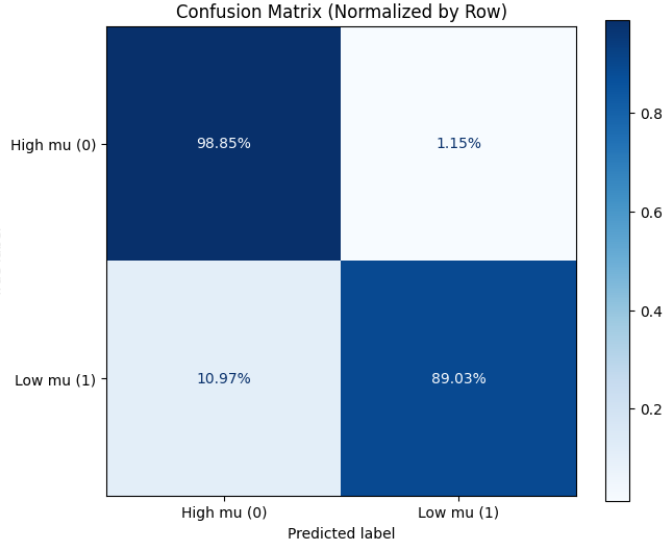}
\caption{Confusion matrix for CNN model.}
\label{fig:cnn_cm}
\end{figure}

\begin{figure}[t]
\centering
\includegraphics[width=\columnwidth]{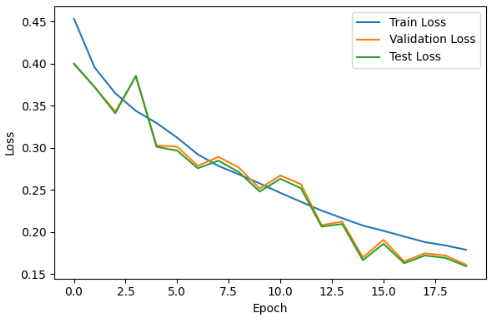}
\caption{CNN Training and validation loss curve over epochs.}
\label{fig:loss_curves}
\end{figure}

Training is performed until 38 epochs after which the train and the validation loss begin to diverge. Fig.~\ref{fig:loss_curves} shows evolution of loss curves on train, validation and test set until 20 epochs. The final test performance is shown in Fig.~\ref{fig:cnn_cm}. The CNN achieves 98.85\% recall for the grip class and 89.03\% recall for the slip class, with grip precision of 90.0\% and slip precision of 98.7\%. Compared to the tree-based model, grip precision is lower (90.0\% vs.\ 92.5\%) and slip recall is also lower (89.03\% vs.\ 92\%), indicating the CNN is more conservative in predicting the slip class. To investigate the reason behind this model behavior, Table~\ref{tab:cnn_importance} shows which channels the CNN model considers the most important in classifying a particular data buffer based on GradCAM \cite{Selvaraju2017} feature interpretability technique used for Deep Learning models. Compared to the tree-based method, the channel importances are much more skewed with preference towards using the acceleration channels. Acceleration from Inertial measurement unit (IMU) typically tends to be noisy due to various reasons such as pavement quality, engine noise, chassis vibrations, etc. One way to mitigate the impact of model overfitting with acceleration data would be to augment input set with smoothed tire force \cite{Zhang2025}, or merely increasing the size of dataset and regularizing the model \cite{Zhao2022} to be robust.

\begin{table}[t]
\caption{1D CNN Channel Importance Scores}
\label{tab:cnn_importance}
\centering
\begin{tabular}{lc}
\toprule
\textbf{Channel} & \textbf{Importance Score} \\
\midrule
Longitudinal Acceleration & 0.2654 \\
Lateral Acceleration & 0.1224 \\
Rear Left Wheel Speed & 0.1027 \\
Rear Right Wheel Speed & 0.0931 \\
Front Left Wheel Speed & 0.0805 \\
Yaw Rate & 0.0717 \\
Vertical Acceleration & 0.0508 \\
Rear Right Wheel Torque & 0.0434 \\
Rear Left Wheel Torque & 0.0405 \\
Front Right Wheel Torque & 0.0386 \\
Front Left Wheel Torque & 0.0337 \\
Steering Wheel Angle & 0.0326 \\
Front Right Wheel Speed & 0.0246 \\
\bottomrule
\end{tabular}
\end{table}

\section{Real World Testing}

\begin{figure*}[t]
\centering
\begin{subfigure}[b]{0.24\textwidth}
\includegraphics[width=\textwidth]{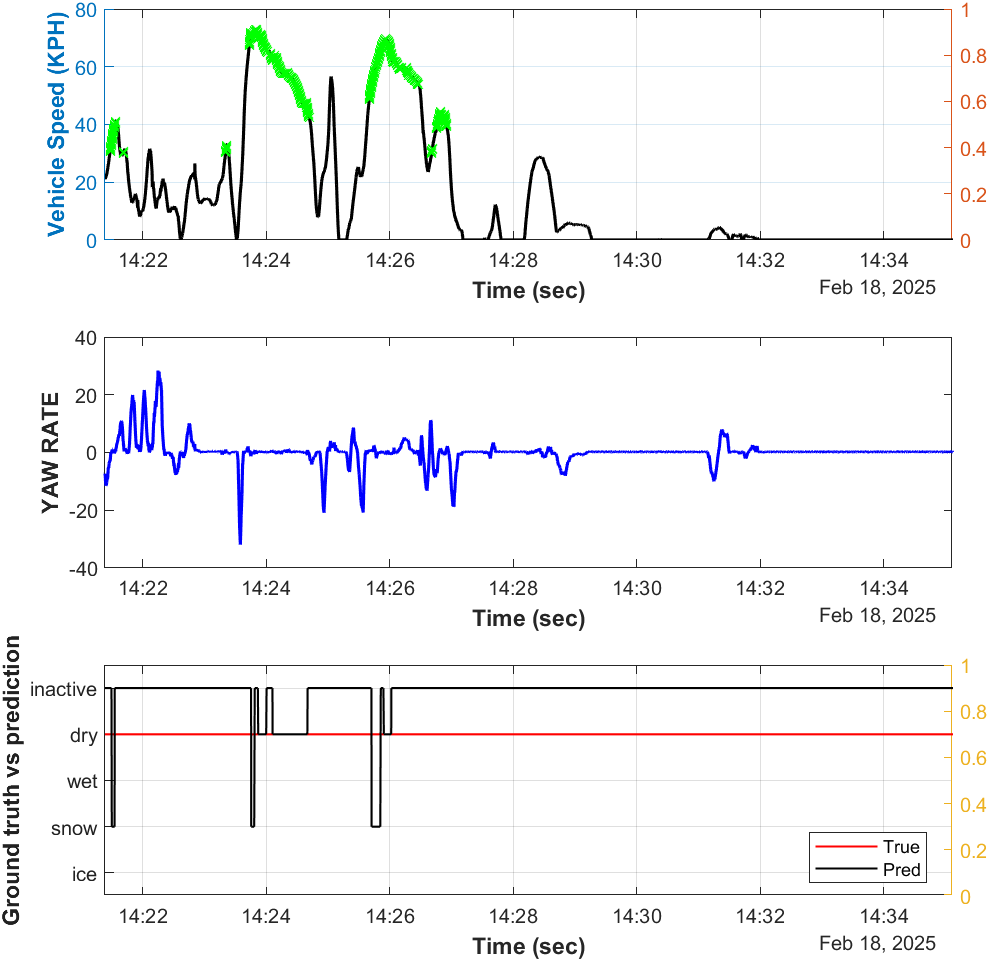}
\end{subfigure}
\hfill
\begin{subfigure}[b]{0.24\textwidth}
\includegraphics[width=\textwidth]{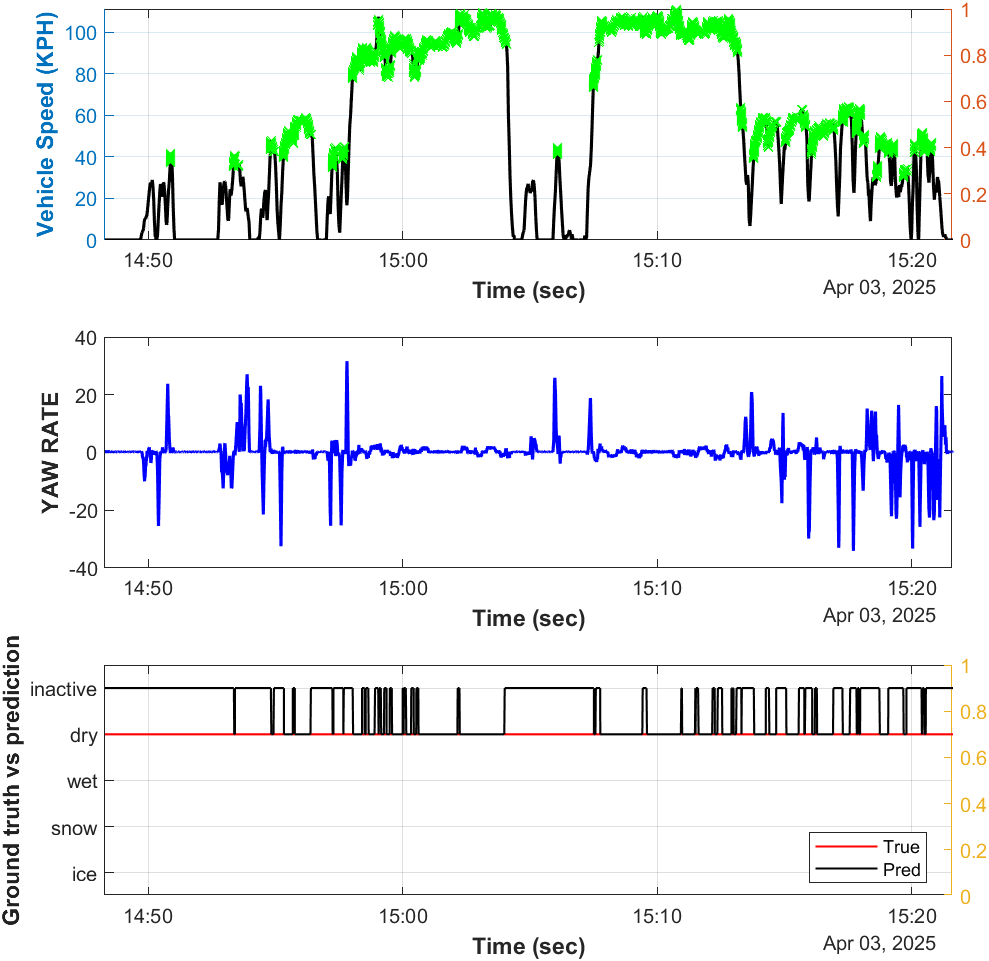}
\end{subfigure}
\hfill
\begin{subfigure}[b]{0.24\textwidth}
\includegraphics[width=\textwidth]{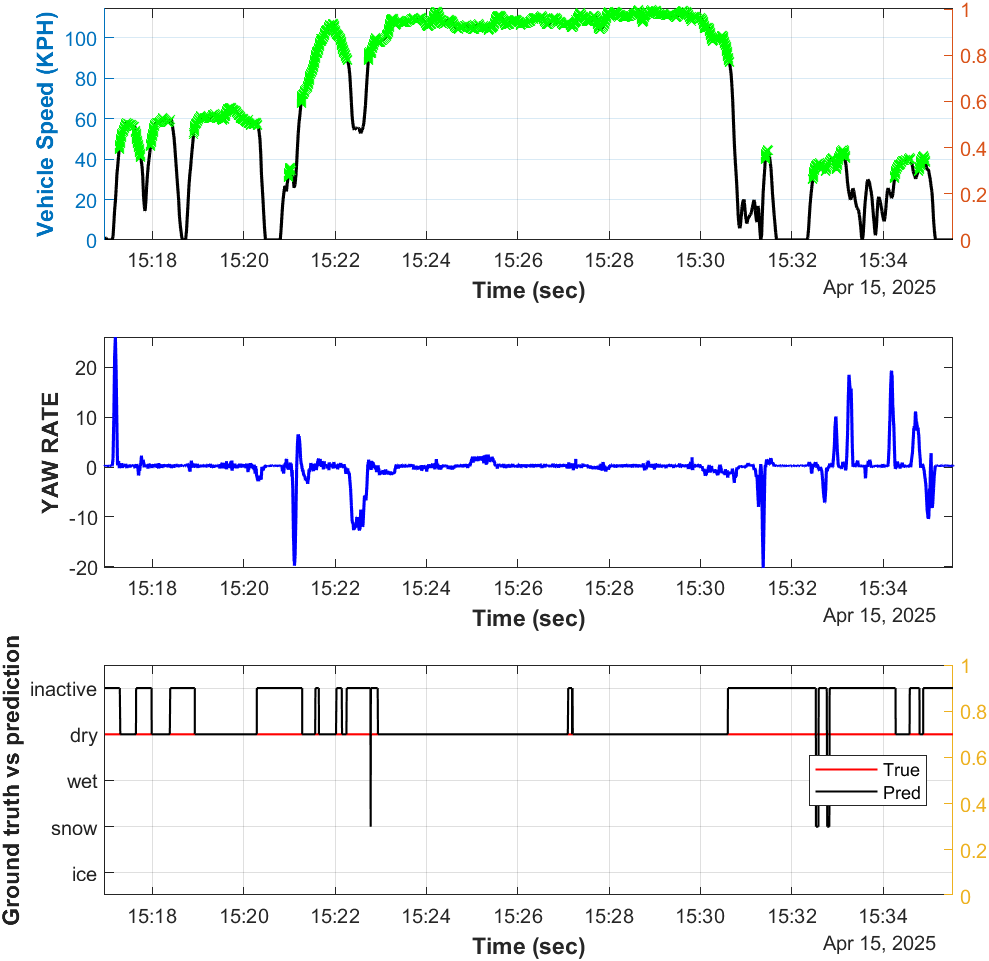}
\end{subfigure}
\hfill
\begin{subfigure}[b]{0.24\textwidth}
\includegraphics[width=\textwidth]{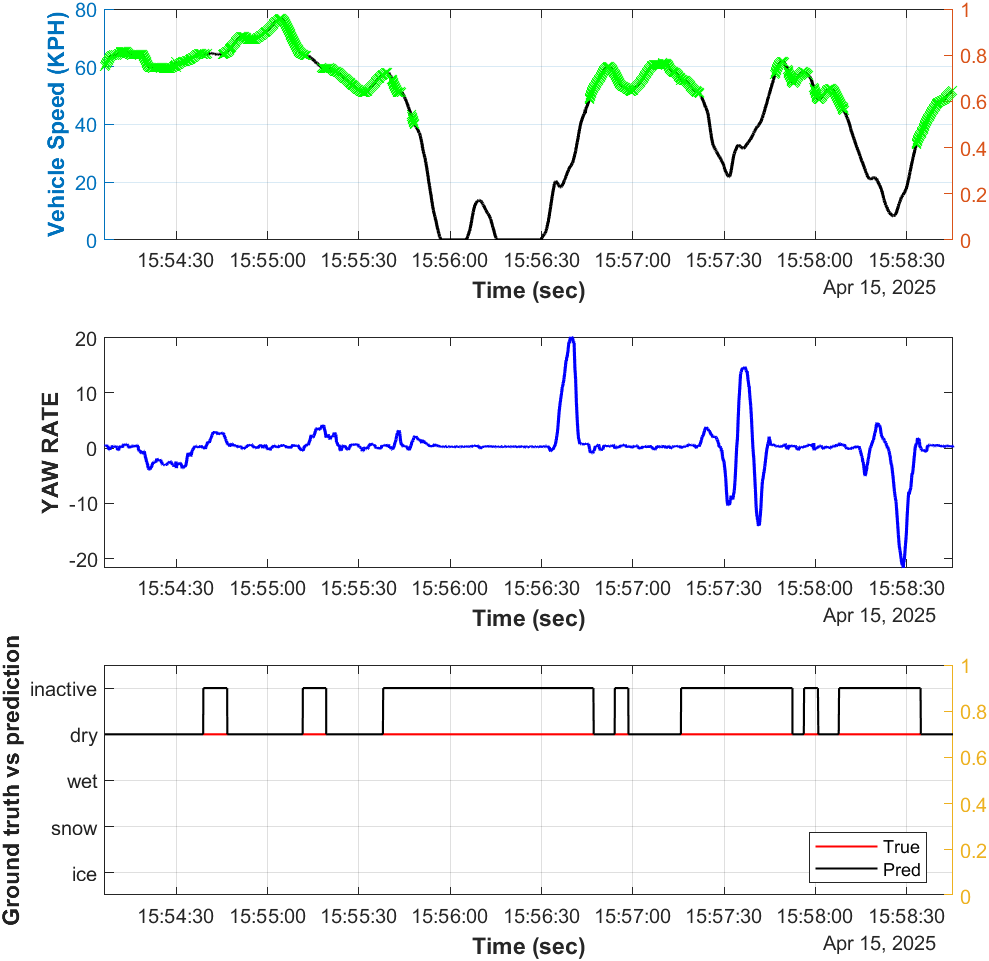}
\end{subfigure}
\caption{Performance of 1D CNN model on dry runs.}
\label{fig:cnn_dry}
\end{figure*}

\begin{figure*}[t]
\centering
\begin{subfigure}[b]{0.24\textwidth}
\includegraphics[width=\textwidth]{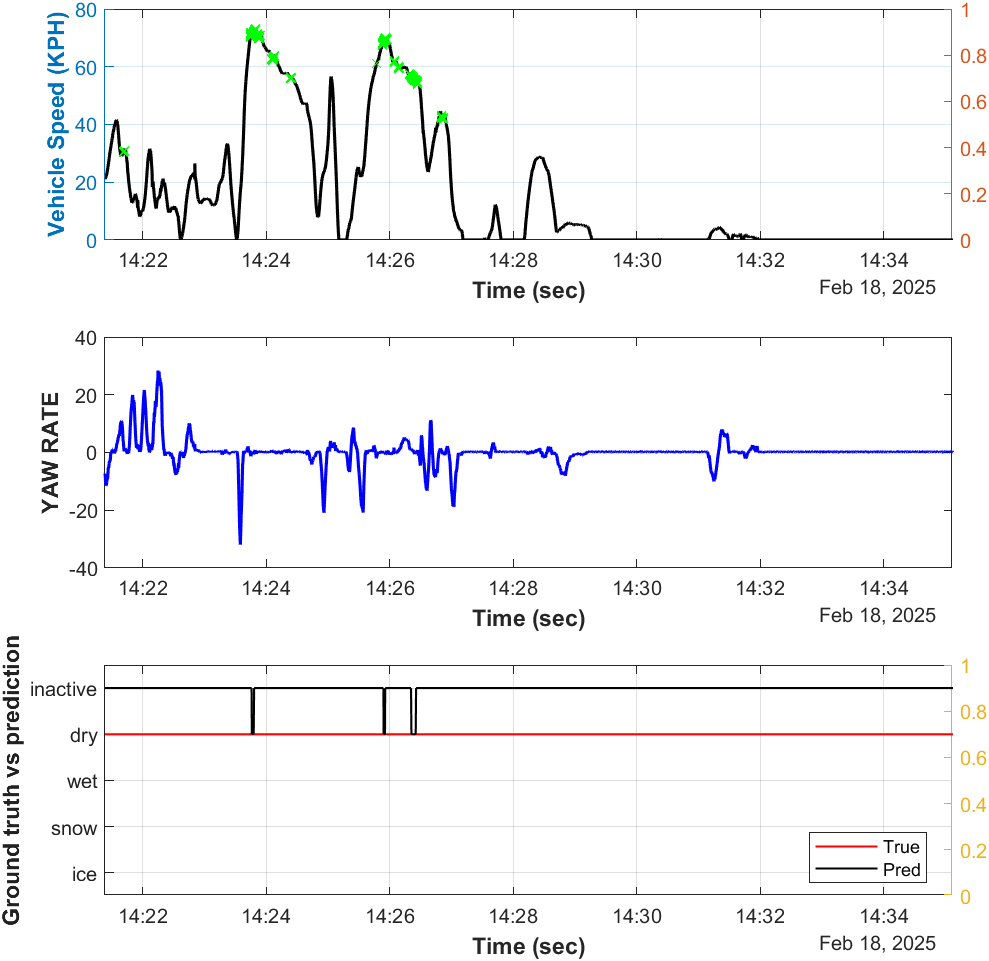}
\end{subfigure}
\hfill
\begin{subfigure}[b]{0.24\textwidth}
\includegraphics[width=\textwidth]{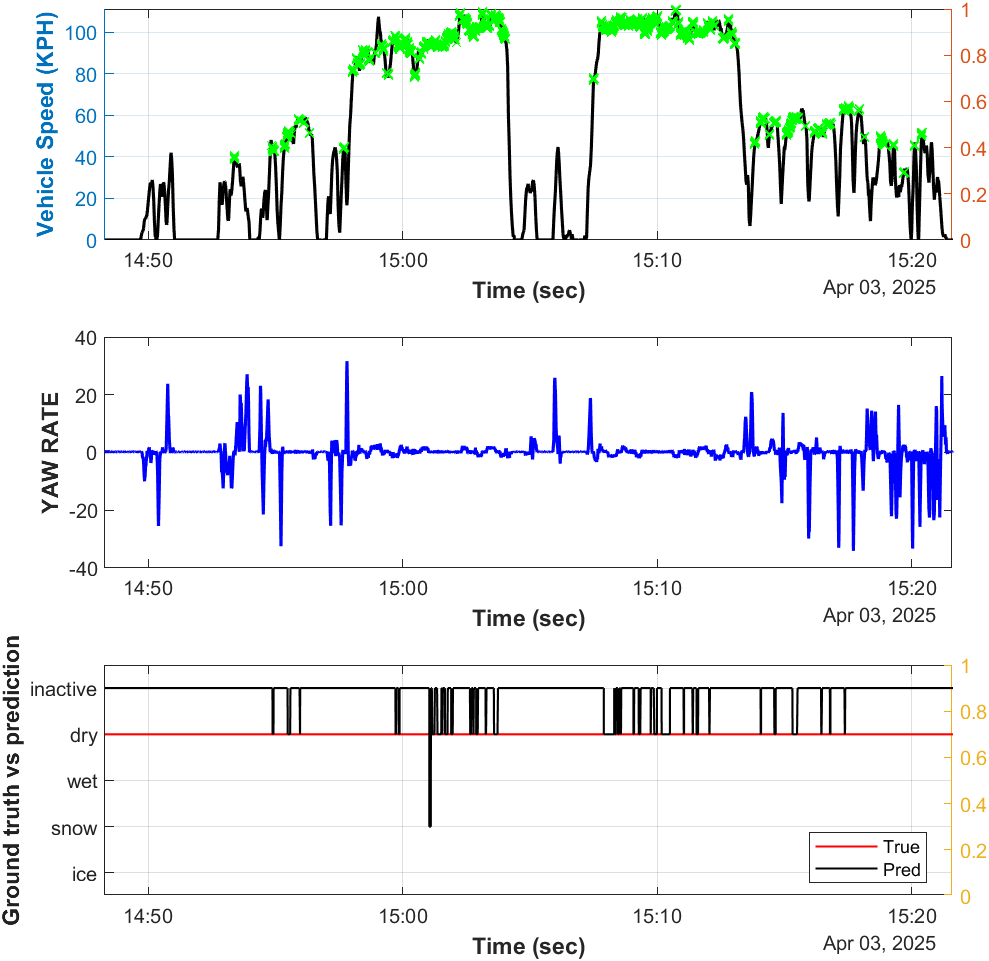}
\end{subfigure}
\hfill
\begin{subfigure}[b]{0.24\textwidth}
\includegraphics[width=\textwidth]{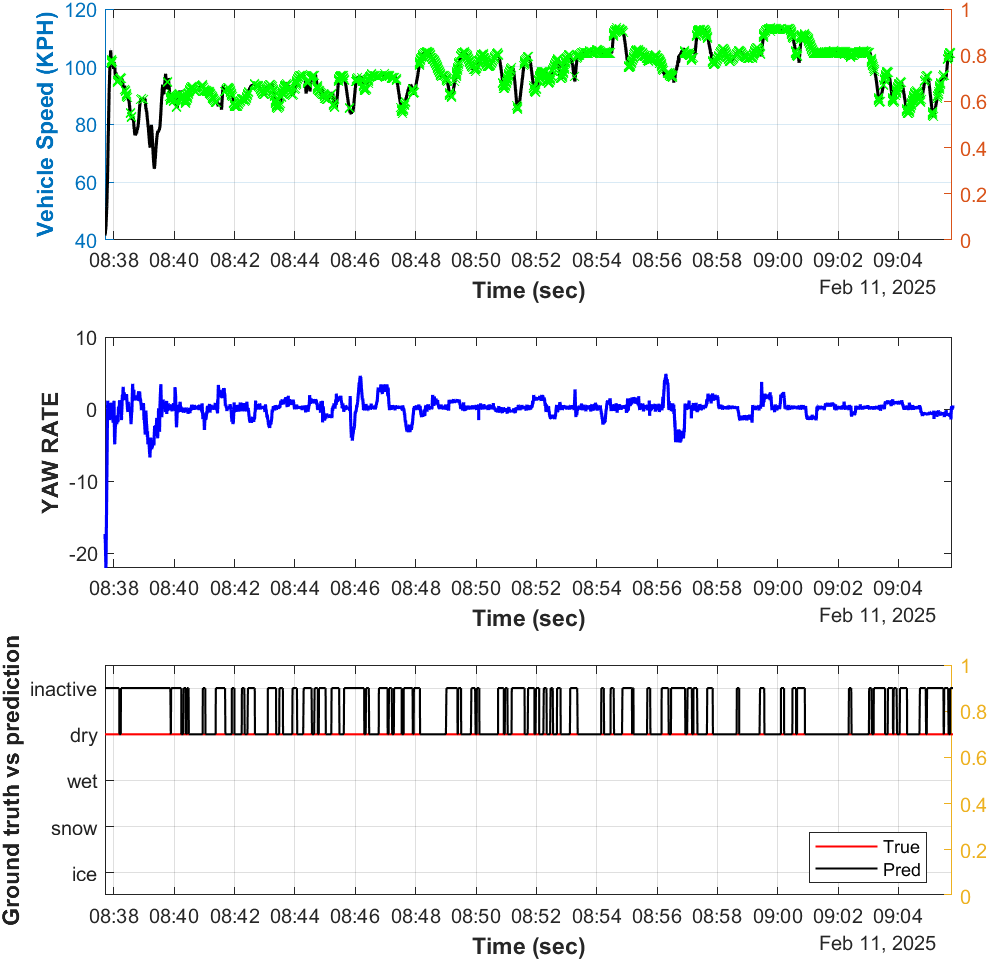}
\end{subfigure}
\hfill
\begin{subfigure}[b]{0.24\textwidth}
\includegraphics[width=\textwidth]{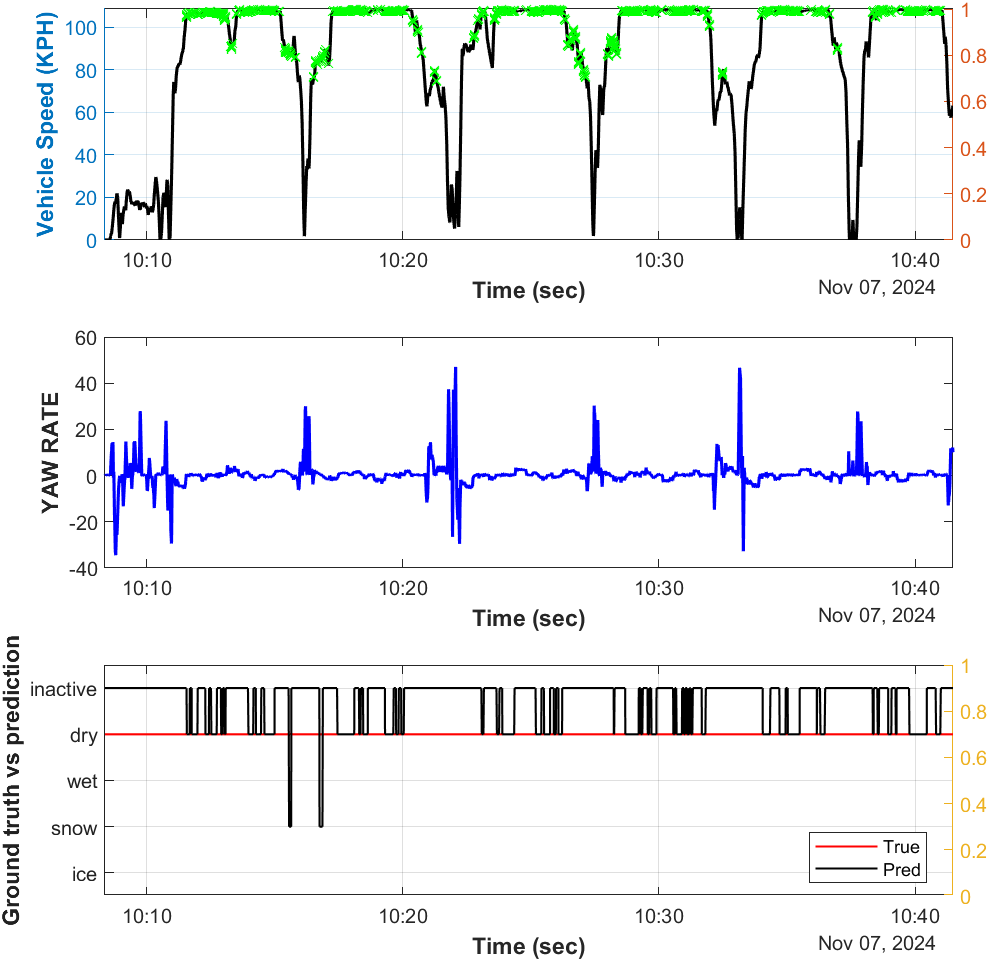}
\end{subfigure}
\caption{Performance of feature based model on dry runs.}
\label{fig:tree_dry}
\end{figure*}

\begin{figure*}[t]
\centering
\begin{subfigure}[b]{0.48\textwidth}
\includegraphics[width=\textwidth]{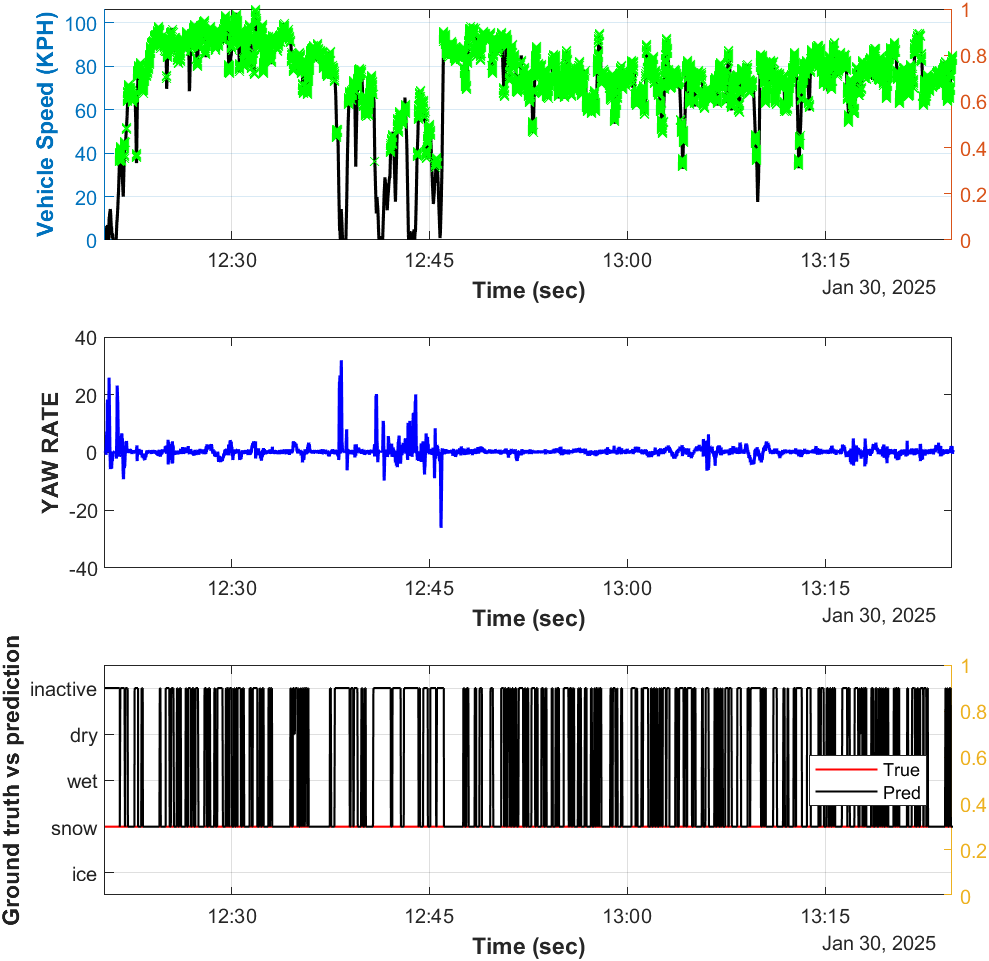}
\end{subfigure}
\hfill
\begin{subfigure}[b]{0.48\textwidth}
\includegraphics[width=\textwidth]{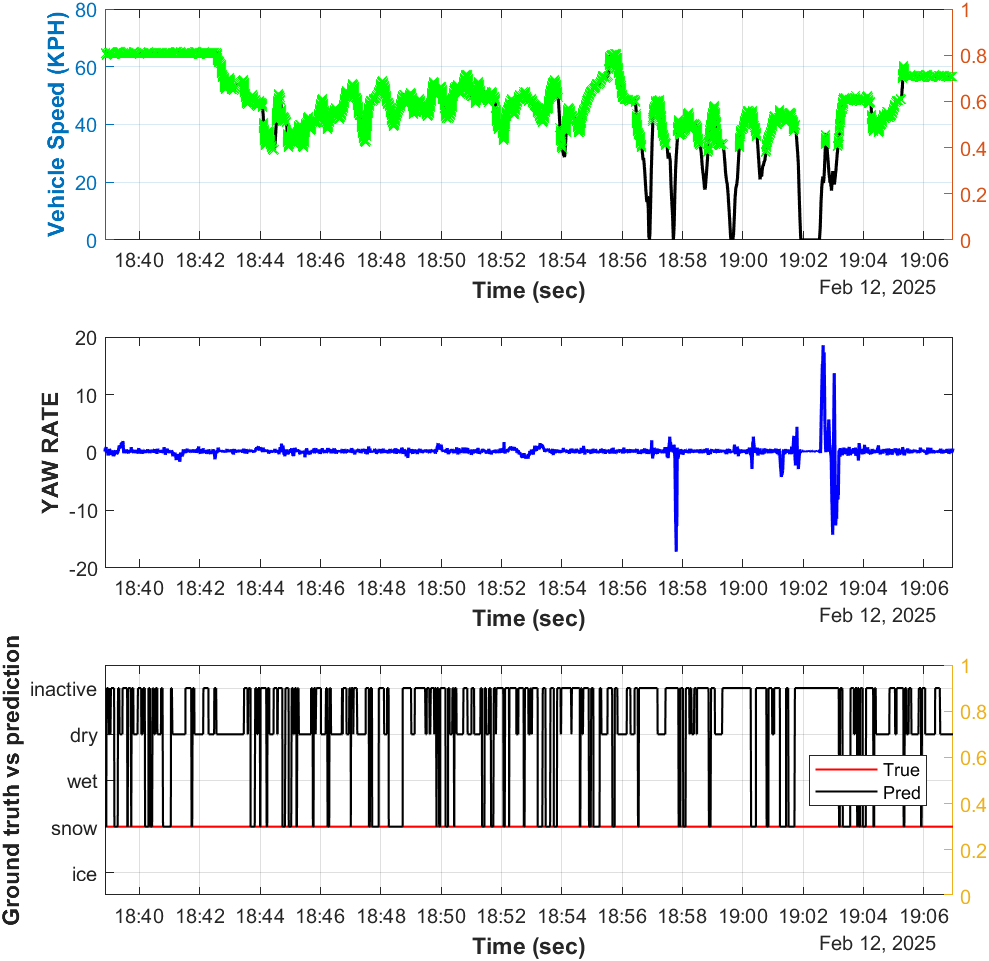}
\end{subfigure}
\caption{Performance of 1D CNN model on snow runs.}
\label{fig:cnn_snow}
\end{figure*}

\begin{figure*}[t]
\centering
\begin{subfigure}[b]{0.48\textwidth}
\includegraphics[width=\textwidth]{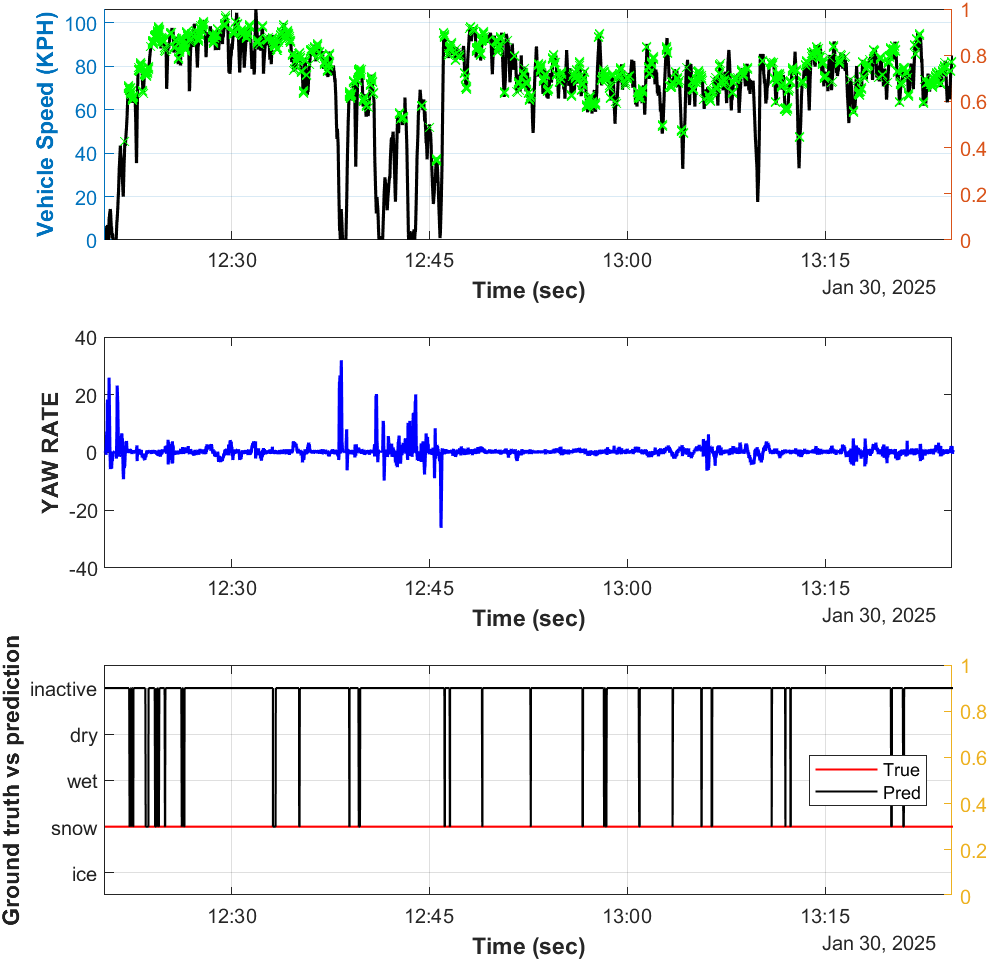}
\end{subfigure}
\hfill
\begin{subfigure}[b]{0.48\textwidth}
\includegraphics[width=\textwidth]{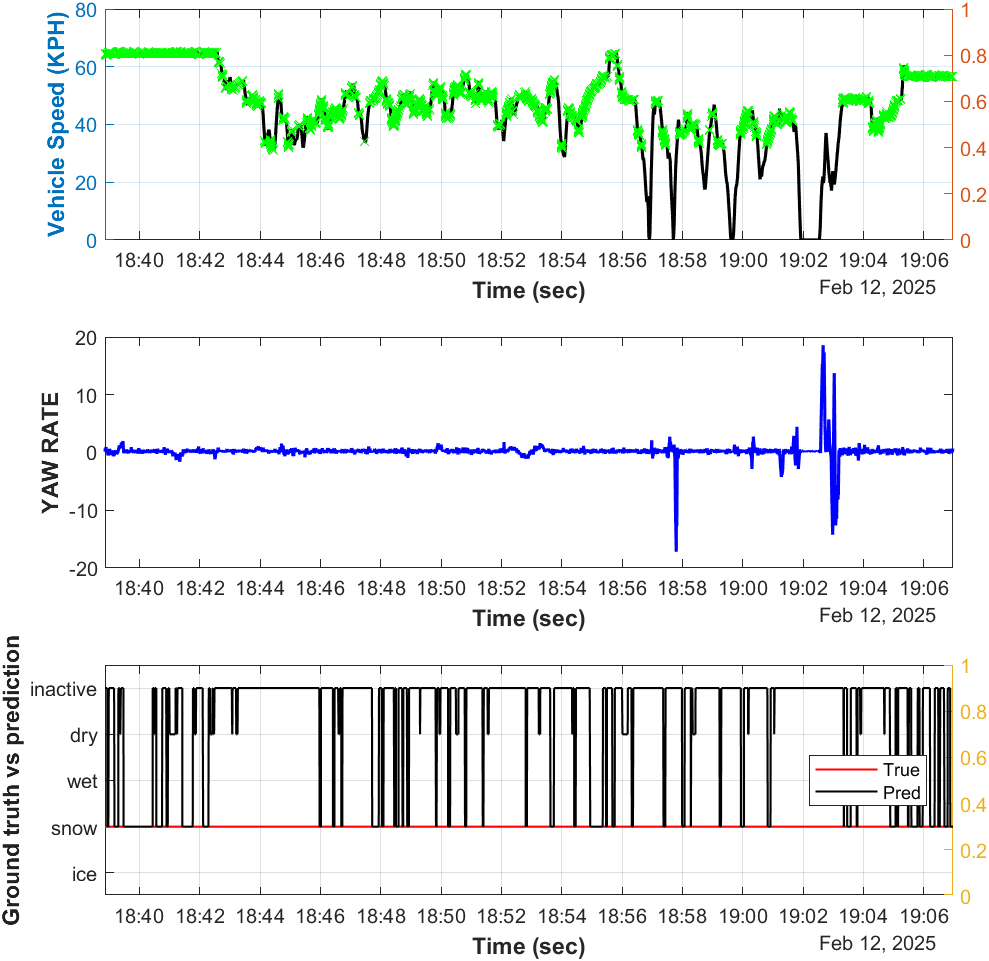}
\end{subfigure}
\caption{Performance of feature based model on snow runs.}
\label{fig:tree_snow}
\end{figure*}

Fig.~\ref{fig:cnn_dry} and Fig.~\ref{fig:cnn_snow} show 1D CNN model performance, Fig.~\ref{fig:tree_dry} and Fig.~\ref{fig:tree_snow} show performance of the tree-based model on a variety of real world maneuvers for longer periods. Both methods exhibit minimal snow false positives on dry surfaces and demonstrate the ability to recall the snow class on snow test runs. The tree-based model shows slightly better performance in detecting the snow class however has a limited operational domain due to how the training was performed. The 1D CNN model is able to operate in a wider range of vehicle operating conditions however shows some sensitivity to report snow false positives during acceleration/braking events due to the acceleration channel overfitting problem addressed in the previous section. Performance on these unseen runs illustrates the potential for model generalization in a real world setting. These test sets are still limited to quantify real world performance metrics. Different driving styles, pavement quality, snow thickness, etc. needs to be tested before certifying the model usable for safety related applications.

\section{Real Time Implementation}

\begin{figure}[t]
\centering
\includegraphics[width=\columnwidth]{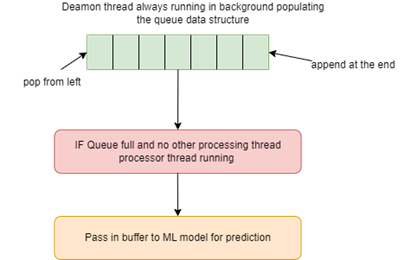}
\caption{Real time inference pipeline.}
\label{fig:rt_pipeline}
\end{figure}

To demonstrate lightweight embedded prototyping feasibility, both algorithms were implemented on Raspberry Pi hardware and were observed to have an inference time under 100 milliseconds. The inference pipeline as shown in Fig.~\ref{fig:rt_pipeline} utilizes two threads: a daemon thread runs in the background, continuously populating points into a queue data structure of the desired window size (4 seconds for the tree-based model and 2 seconds for the 1D CNN), while a processor thread performs inference on the buffered data.

To ensure stable behavior in the outputs, an additional circular buffer of fixed size 30 is maintained where the results are populated. A smoothing logic as shown in Algorithm~\ref{alg:modefilter} is used to display a stable output on a display monitor and/or send the outputs to a prototyping unit such as MicroAutobox. This would mean the detection delay can be expected to be around 3 seconds.

\begin{algorithm}[t]
\caption{Mode Filter with Thresholding for Classification Smoothing}
\label{alg:modefilter}
\begin{algorithmic}[1]
\REQUIRE $\text{data}$ -- array of classification values (0=Dry, 1=Snow, 4=Inactive)
\REQUIRE $\text{window\_size}$ -- size of circular buffer
\ENSURE $\text{filtered\_data}$ -- smoothed classification array
\STATE \textbf{function} ModeFilter(data, window\_size)
\IF{window\_size mod $2 = 0$}
    \STATE raise ValueError(``Window size must be odd.'')
\ENDIF
\STATE $\text{half\_window} \leftarrow \lfloor \text{window\_size} / 2 \rfloor$
\STATE $\text{filtered\_data} \leftarrow \text{copy(data)}$
\FOR{$i \leftarrow 0$ \TO $\text{length(data)} - 1$}
    \STATE $\text{start\_index} \leftarrow \max(0, i - \text{half\_window})$
    \STATE $\text{end\_index} \leftarrow \min(\text{length(data)}, i + \text{half\_window} + 1)$
    \STATE $\text{window} \leftarrow \text{data}[\text{start\_index} : \text{end\_index}]$
    \STATE Count occurrences of each value in window
    \STATE $\text{threshold} \leftarrow 0.8 \times \text{length(window)}$
    \IF{$\text{count}(1) \geq \text{threshold}$}
        \STATE $\text{filtered\_data}[i] \leftarrow 1$ \COMMENT{Snow}
    \ELSIF{$\text{count}(0) \geq \text{threshold}$}
        \STATE $\text{filtered\_data}[i] \leftarrow 0$ \COMMENT{Dry}
    \ELSE
        \STATE $\text{filtered\_data}[i] \leftarrow 4$ \COMMENT{Inactive}
    \ENDIF
\ENDFOR
\RETURN filtered\_data
\STATE \textbf{end function}
\end{algorithmic}
\end{algorithm}

\section{Interpretability}

\begin{figure}[t]
\centering
\includegraphics[width=\columnwidth]{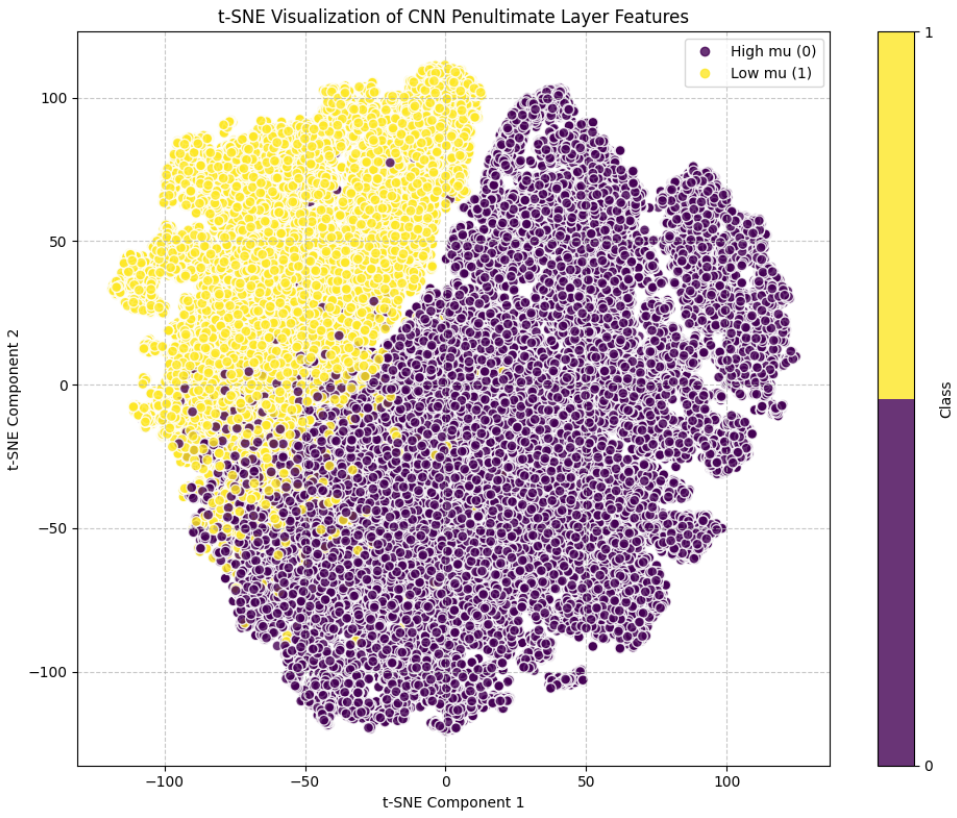}
\caption{t-SNE embeddings of CNN penultimate features.}
\label{fig:tsne}
\end{figure}

Machine Learning models work well in the realm of training data distribution, but can struggle on out of distribution evaluation. It is possible that the performance metrics are good for a model, but it has learned a non-physical phenomenon that has worked well in that data distribution but is not physical enough to generalize. In order to avoid models having this behavior, it is crucial to conduct interpretability studies. Fig.~\ref{fig:tsne} shows the high dimensional penultimate layer features projected onto 2 dimensions using t-SNE. A clean-cut separation between the 2 classes shows that the neural network has been able to take the input buffers and project them onto a higher dimensional space where there is good separability.

\section{Conclusion}

In this work, it is demonstrated that data driven methods can detect slippery conditions solely using onboard signals. However, the drawback of these methods is their sensitivity to other signatures present in the signals due to pavement quality, or aggressive driver behavior. To mitigate these challenges, it is recommended to have multiple data driven methods based on different modalities such as image \cite{Liu2022, Cordes2022, Zhao2024} and weather data \cite{DiLorenzo2023} as well as a physics based method based on onboard signals report the road slipperiness condition and their agreement or disagreement can be used to arrive at confidence scores. Compared to traditional methods, data-driven models offer the potential to scale across different vehicle and tire platforms and run efficiently. This is particularly advantageous as multiple algorithms typically run on a vehicle's Electronic Control Unit (ECU), and no single algorithm can dominate resources. A deep learning method should be able to sufficiently capture the pattern differences provided the scale and quality of data is large enough and feature based methods can provide insights into pruning the neural network model. It is also important to quantify model performance in different driving conditions and pavement quality before moving towards production grade algorithm. Several limitations should be noted. First, this work addresses binary classification (grip vs.\ slip) only, grouping dry and damp surfaces as grip and snow and ice as slip. Wet, slush, and surface transitions are excluded, which substantially reduces the difficulty of the classification problem compared to full road-surface state estimation. Second, the evaluation on unseen drives is qualitative; per-run false-positive rates, detection delay, and robustness across vehicle speeds, acceleration levels, and pavement types are not quantified. Third, the train/test split strategy and potential temporal overlap of sliding windows should be examined more rigorously to rule out data leakage. Finally, this work does not address production integration aspects or ASPICE compliance.

This work is an early step toward demonstrating the feasibility of data-driven methods for detecting slipperiness during low tire excitation using only production-feasible onboard signals. Future work will extend to multi-class classification including wet and slush conditions, quantify real-time performance metrics such as detection delay in jump-mu and split-mu \cite{Dong2017} test scenarios, and evaluate robustness across diverse driving conditions and vehicle platforms.

\bibliography{ref}

\end{document}